\pdfoutput=1

\documentclass[11pt]{article}

\usepackage{naacl2021}

\usepackage{times}
\usepackage{latexsym}

\usepackage[T1]{fontenc}

\usepackage[utf8]{inputenc}

\usepackage{microtype}

\usepackage{booktabs}
\usepackage{tabularx}

\title{No Intruder, no Validity: Evaluation Criteria for\\Privacy-Preserving Text Anonymization}
\author{Maximilian Mozes$^{1,2,3}$ \qquad Bennett Kleinberg$^{4,1}$\\
$^1$Department of Security and Crime Science, University College London\\
$^2$Dawes Centre for Future Crime, University College London\\
$^3$Department of Computer Science, University College London\\
$^4$Department of Methodology and Statistics, Tilburg University\\

\small{\texttt{maximilian.mozes@ucl.ac.uk}\qquad\texttt{bennett.kleinberg@tilburguniversity.edu}}}
\date{}

\begin{document}
\maketitle
\begin{abstract}
For sensitive text data to be shared among NLP researchers and practitioners, shared documents need to comply with data protection and privacy laws. There is hence a growing interest in automated approaches for text anonymization. However, measuring such methods' performance is challenging: missing a single identifying attribute can reveal an individual's identity. In this paper, we draw attention to this problem and argue that researchers and practitioners developing automated text anonymization systems should carefully assess whether their evaluation methods truly reflect the system's ability to protect individuals from being re-identified. We then propose \textsc{TILD}, a set of evaluation criteria that comprises an anonymization method's technical performance, the information loss resulting from its anonymization, and the human ability to de-anonymize redacted documents. These criteria may facilitate progress towards a standardized way for measuring anonymization performance.

\end{abstract}

\section{Introduction}

Vast developments in natural language processing have led to an increased interest in the application of text processing approaches across fields, including the biomedical~\cite{armengol2019pharmaconer}, law enforcement~\cite{yang2014authorship} and business sectors~\cite{davenport2018analytics}. Part of the drive for applying methods beyond research contexts is the enormous availability and increasing relevance of digital text data as a rich source of information. Textual corpora can now be analyzed with quantitative methods, for example, to extract information and harness methods from machine learning for predictive tasks (e.g., text classification and natural language generation). However, the great potential of text data for researchers, businesses and policymakers, rests on a key precondition: access to and the sharing of text data.

Researchers could further push NLP boundaries and set out to address topical problems if data are more easily shareable. Similarly, businesses would benefit as data flow becomes less constrained (e.g., by providing access to real-world data in tender procedures). Lastly, cross-sector relationships, where academic research feeds into policy-making or business processes---or vice versa---rely on access to text data. Especially in contexts where the text data are sensitive by nature (e.g., police reports, health care dossiers) or protected by privacy laws (e.g., the General Data Protection Regulation; GDPR), data sharing is currently either non-existent or requires extensive data-sharing agreement negotiations. Consequently, academic research often resorts to easy-to-get data that is removed from an actual problem context, and public and private sector organizations who possess sensitive data only rarely harness the powerful methods developed at academic institutions.

\begin{table*}[t!]
\centering
\begin{tabularx}{\textwidth}{X>{\hsize=.3\hsize}X}
\toprule
\multicolumn{1}{c}{\textbf{Example statement}} & \multicolumn{1}{c}{\textbf{Explanation}} \\ \midrule
\textit{PERSON1, nicknamed the \color{red}\textbf{“Iron Lady”}\color{black}, served as LOCATION1 OCCUPATION1 from DATE1 to DATE2.} & \textit{"Iron Lady"} likely to reveal \textit{Margaret Thatcher's} identity \\ \midrule
\textit{OTHER1 produced and recorded a variety of musical hits such as OTHER2, \color{red}{\textbf{“Bohemian Rhapsody”}}\color{black}\, and OTHER3.} & \textit{“Bohemian Rhapsody”} likely to reveal \textit{Queen's} identity \\ \midrule
\textit{PERSON1 known for PRONOUN role as PERSON2 struggles to say the word penguin.}& The penguin reference may reveal \textit{Benedict Cumberbatch}\footnotemark[1] \\
\bottomrule
\end{tabularx}
\caption{Illustration of failure cases in the anonymization process where missing a single personally-identifying attribute invalidates the success of the text anonymization.}
\label{tab:anon-failure-cases}
\end{table*}

The common solution to reconcile privacy protection with the sharing of text data is text anonymization. While the idea of redacting documents is not new, the volume of the data now available and needed by many NLP approaches makes manual efforts futile. A solution is automated text anonymization which, broadly, aims to build systems that automatically redact personally-identifying information from text, to make the anonymized text fit for sharing without violating the privacy of individuals identifiable from the text. In this paper, we \textit{draw attention to the problems of current evaluation standards} of text anonymization systems and propose \textsc{TILD}, a \textit{set of evaluation criteria that we recommend any suchlike system to be tested against}. 

\section{Requirements for text anonymization}

Text anonymization can serve two purposes, each of which determines the sophistication of the anonymization approach. First, one may decide to fully redact documents by removing any trace of personally-identifying information. Such removal of information may apply to government documents released to the public without any secondary analysis envisioned. The second purpose is the sharing of text data to enable secondary analysis based on the shared documents. While these two purposes both share the aim of removing personally-identifying information, the critical difference is what the original information (e.g., names, locations, attributes) are replaced with. Simply crossing out the relevant information might render the documents useless for secondary analyses. One of the criteria we outline below postulates that the difference in findings based on the original and anonymized documents should be minimal. Current anonymization systems lack a framework that can be used for evaluation, making it difficult to compare approaches and leaves an ambiguous picture for stakeholders and data owners deciding which anonymization system works best for their purpose.

\footnotetext[1]{\url{https://www.vanityfair.com/hollywood/2014/09/benedict-cumberbatch-penguins}}

\begin{table*}[t!]
\begin{tabularx}{\textwidth}{lXX} 
\toprule
\multicolumn{1}{c}{\textbf{Criterion}} & \multicolumn{1}{c}{\textbf{Description}} & \multicolumn{1}{c}{\textbf{How to test it?}} \\ \midrule
\textit{Technical tool evaluation} & Measures how well the system detects relevant information & Reporting of common performance metrics (detection accuracy, precision) \\ \midrule
\textit{Information loss} & Reports the difference in prediction tasks (utility loss) or linguistic variables (construct loss) between original and anonymized texts & Statistical testing of the loss deviation from a tolerance value \\ \midrule
\textit{De-anonymization} & Assesses whether a motivated intruder can re-identify individuals in anonymized text & Pentesting, game-like studies with human participants tasked to identify individuals \\
\bottomrule
\end{tabularx}
\caption{Overview of the three \textsc{TILD} evaluation criteria \textit{technical tool evaluation}, \textit{information loss} and \textit{de-anonymization}.}
\label{tab:anon-criteria}
\end{table*}

\subsection{Evaluation problems}
Recently, several works dealing with text anonymization for multiple languages have been put forward~\cite[e.g.,][]{7743936, kleinberg2017web, 10.1007/978-3-030-00063-9_13, 10.1007/978-3-030-00202-2_24, adams-etal-2019-anonymate,berg-dalianis-2019-augmenting,francopoulo2020anonymization}. For example,~\citet{adams-etal-2019-anonymate} proposed an automated text anonymization system based on named entity recognition and coreference solution modules. The proposed method identifies sensitive phrases in a document and replaces them with generic symbols or entity-specific replacement tokens (e.g., \textit{LOCATION} or \textit{ENGLISH\_CITY} for \textit{London}).~\citet{francopoulo2020anonymization} developed an anonymization system in the context of Customer Relationship Management consisting of a named entity recognition model, an entity linker and a substitution method. While these approaches include a learning-based component for the automated anonymization of text data,~\citet{strathern2020qualanon} presented a tool that assists practitioners in manually anonymizing texts. Some other tools are available commercially, but these are typically not evaluated against a peer-reviewed standard.

Existing works to automated text anonymization~\cite[e.g.,][]{adams-etal-2019-anonymate,francopoulo2020anonymization} predominantly use fully automated precision- and recall-based methods to assess the performance of a system. While such an evaluation approach measures a system's ability to extract relevant information from text, we here argue that it is not sufficient to assess a text anonymization system's performance. Our argumentation against the use of purely automated methods for evaluating a text anonymization system is twofold. 

First, missing a single but critical attribute in a text invalidates any other performance or evaluation metric that explains which fraction of the required entities have been identified and anonymized. This is illustrated in Table~\ref{tab:anon-failure-cases}. Here, an anonymization method correctly redacts all but one of the personally-identifying attributes in a given text and would score high on purely quantitative methods. Yet missing a single detail would reveal the individual's or group's identity.

Second, while identifying and redacting personally-identifying information is a necessary criterion for successful anonymization, it is not a sufficient one. For anonymized texts to be useful for data analysis purposes, it needs to be ensured that anonymization methods preserve the context of a text to the degree that the transformed text retains its usefulness for text mining tasks (e.g., sentiment analysis and topic modelling). We refer to this below as the \textit{preservation of information}~\cite[see][]{feyisetan2020privacy}.

\section{The \textsc{TILD} evaluation criteria}
While some attention has been paid to the development of technical solutions to the anonymization problem, little attention has been given to how we can assess the success of text anonymization efforts. That evaluation question is the most fundamental one from a data protection point of view and thus the central aspect which decision-makers need to ask. What is currently used as an evaluation approach falls short of the fundamental aim of text anonymization. 

To address this problem, we propose three criteria against which any anonymization approach should be evaluated. We denote this set of criteria with \textsc{TILD}, comprising a model's \underline{\textbf{t}}echnical tool evaluation, the \underline{\textbf{i}}nformation \underline{\textbf{l}}oss resulting from the anonymization, and the human ability to \underline{\textbf{d}}e-anonymize the redacted documents (see Table~\ref{tab:anon-criteria}). We also illustrate how \textsc{TILD} can be put into practice to evaluate anonymization systems. The requirements should guide research efforts to pay attention to these aspects and equip decision-makers (academic, business, policymaking) with a set of criteria to look out for.

Of the three criteria comprised by \textsc{TILD}, two are currently somewhat present in text anonymization work (i.e., the technical evaluation and the preservation of information). The third criterion (person re-identification) has thus far been ignored but sits at the heart of what anonymization is supposed to do.

\subsection{Technical tool evaluation}
A simple technical tool evaluation is frequently contained in assessments of a system's performance in detecting (tagging) relevant information in text data (e.g., the accuracy or the precision). For example, a system might be evaluated by the percentage of locations, persons and dates it detects in a text. 
For anonymization systems, meeting this criterion is necessary but by no means sufficient: it makes no statement about how much information is lost during anonymization, and it inherently assumes that removing specific pieces of information does indeed render persons unidentifiable from the text. Thus, the technical evaluation cannot be used as a sole criterion for the goodness of anonymization efforts.

\subsection{Information loss}
A system that performs well on the technical evaluation level can be assessed on the next level: preserving utility. This aspect is critical for secondary data analyses conducted on the anonymized text data. Information loss is defined broadly here as the difference in some quantitative measure between the original text and its anonymized version. Information loss can be assessed statistically by testing whether the difference between original vs anonymized text is markedly larger than can be tolerated (or strictly: larger than zero). We differentiate between two kinds of information loss.

\paragraph{Utility loss.}The utility loss~\cite{feyisetan2020privacy} has been used to assess the difference in performance (e.g., on a benchmark sentiment classification task) that is obtained when using the original texts versus the anonymized texts. High utility loss implies that the performance differs so that the anonymization distorts any secondary analysis results.

\paragraph{Construct loss.}The construct loss is not focused on performance differences. Instead, it measures the differences between original and anonymized text on some higher-order linguistic measure (e.g., those derived with dictionary tools frequently used in socio- and psycholinguistic research). High construct loss is undesirable because it would indicate that the anonymization procedure affected the text to the extent that findings from the original do not replicate in the anonymized one.
Both information loss criteria address divergent results in anonymized texts compared to original ones, but they alone are yet insufficient for the core anonymization purpose.

\subsection{De-anonymization}
The core purpose of text anonymization is to protect the privacy of individuals. Consequently, any text anonymization effort should be tested about its ability to not reveal the identity of individuals. Since an individual can be identified by more than just their name, address or age (see Table~\ref{tab:anon-failure-cases}), it may not suffice to test whether directly identifying and protected attributes are leaking. Purely automated evaluation metrics (i.e., technical and information loss criteria) are, therefore, insufficient for an assessment of an anonymization system. 

Instead, a more viable approach is de-anonymization penetration testing by focusing on two possible outcomes: successful or failed anonymization. Such an evaluation can be realized via a \textit{motivated intruder test}~\cite{information2012anonymisation}. The UK's Information Commissioner's Office suggests evaluating the quality of any anonymization system (i.e., not just text) by tasking an individual human being (the intruder) with the de-anonymization of any persons from an anonymized piece of information. Such a test puts a human intruder---who is allowed and encouraged to use external resources---into a game-like scenario with the goal of re-identifying individuals. Crucially, the intruder does not have privileged information, nor do they have any specialist knowledge beyond what is available publicly. This provides a more realistic demand on the anonymization because it excludes scenarios with highly privileged and hence directly revealing information (e.g., a physician knowing that a condition \textit{X} only occurs in patient \textit{A}).
If an intruder can de-anonymize a text, the anonymization failed.

\paragraph{Anonymization intrusion in practice.}A possible way of putting the de-anonymization criterion of the proposed \textsc{TILD} guidelines to the test would be via online crowdsourcing. Potential intruders could be recruited on crowdsourcing platforms such as Amazon's MTurk or Prolific Academic. In an online behavioral study, participants would be asked to read an anonymized piece of information and then tasked with re-identifying any individual mentioned or described in the text. The intruder should be encouraged (and potentially even incentivized) to use any external resources available to them (such as a web search engine). The ratio of successful de-anonymizations (i.e., where intruders correctly revealed the identities of individuals) to all attempts serves as the key evaluation metric for a text anonymization system.

\section{Conclusion}
In this paper, we argued that relying on a technical evaluation is insufficient to measure the true capabilities of automated text anonymization systems. We proposed \textsc{TILD}, a set of more comprehensive evaluation criteria. In addition to technical evaluation metrics, \textsc{TILD} measures an anonymization system's ability to preserve task-relevant linguistic concepts and the degree to which it truly preserves the privacy of individuals mentioned in a text. We encourage researchers and practitioners to adopt more diverse evaluation criteria for anonymization systems to provide an assessment that more rigorously reflects the potential of their proposed methods.

\section*{Acknowledgements}
We would like to thank Toby Davies for helpful discussions about this work.

\bibliography{ref}
\bibliographystyle{acl_natbib}

\end{document}